%% file: paper.tex
\documentclass[10pt,twocolumn,letterpaper]{article}

\usepackage{wacv}
\usepackage{times}
\usepackage{epsfig}
\usepackage{graphicx}
\usepackage{amsmath}
\usepackage{amssymb}

\usepackage{booktabs}
\usepackage{algpseudocode}
\usepackage{algorithm}
\usepackage{mathtools}
\usepackage{color}
\usepackage{multirow,graphicx,paralist}
\usepackage{ colortbl}
\usepackage{ tabularx}
\definecolor{Gray}{gray}{0.85}
\definecolor{LightCyan}{rgb}{0.88,1,1}
\definecolor{antiquefuchsia}{rgb}{0.57, 0.36, 0.51}
\definecolor{bleudefrance}{rgb}{0.19, 0.55, 0.91}
\usepackage[dvipsnames]{xcolor}
\usepackage{csquotes}

\newcommand\Tstrut{\rule{0pt}{1.7ex}}

\usepackage{upquote}

\usepackage[pagebackref=true,breaklinks=true,letterpaper=true,colorlinks,bookmarks=false]{hyperref}

\wacvfinalcopy 

\def\wacvPaperID{263} 

\ifwacvfinal\fi
\begin{document}
	
\title{Distributed Iterative Gating Networks for Semantic Segmentation}

\author{Rezaul Karim$^{1}$\hspace{0.16cm}, Md Amirul Islam$^{2,3}$\hspace{0.16cm}, Neil D. B. Bruce$^{2,3}$ \\
	$^1$York University, $^2$Ryerson University, $^3$Vector  Institute  for  Artificial  Intelligence\\
	{\tt\small karimr31@yorku.ca, amirul@scs.ryerson.ca, bruce@ryerson.ca}
}

\maketitle

\begin{abstract}
 In this paper, we present a canonical structure for controlling information flow in neural networks with an efficient feedback routing mechanism based on a strategy of Distributed Iterative Gating (DIGNet). The structure of this mechanism derives from a strong conceptual foundation, and presents a light-weight mechanism for adaptive control of computation similar to recurrent convolutional neural networks by integrating feedback signals with a feed forward architecture. In contrast to other RNN formulations, DIGNet generates feedback signals in a cascaded manner that implicitly carries information from all the layers above. This cascaded feedback propagation by means of the propagator gates is found to be more effective compared to other feedback mechanisms that use feedback from output of either the corresponding stage or from the previous stage. Experiments reveal the high degree of capability that this recurrent approach with cascaded feedback presents over feed-forward baselines and other recurrent models for pixel-wise labeling problems on three challenging datasets, PASCAL VOC 2012, COCO-Stuff, and ADE20K.
 \end{abstract}

\input introduction.tex
\input background.tex

\input approach.tex

\input experiments.tex

\input conclude.tex

{\small
\bibliographystyle{ieee}
\bibliography{paper}
}

\end{document}

%% file: introduction.tex
\section{Introduction}\label{sec:intro}
Deep learning models have achieved a high degree of success for problems involving dense pixel labeling~\cite{long15_cvpr,chen15_iclr,noh15_iccv,badrinarayanan15_arxiv,yu2015multi,ghiasi2016laplacian,Islam_2017_CVPR,refinenet,chen2018deeplab} with a wide range of associated applications\cite{li2016iterative,liang2015convolutional}. Improvements in this domain have come by virtue of increasingly deep networks~\cite{krizhevsky12_nips,simonyan15_iclr,szegedy2015going,he2016deep}, pre-training that leverages data from multiple large scale datasets~\cite{deng09_cvpr,lin14_eccv} to boost overall performance, and innovations on architectural properties of networks. In this paper, we focus heavily on the last of these categories in proposing a scheme for efficient selection and routing of feed-forward information in neural networks.

There are a few specific considerations that motivate this paper, which presents a simple lightweight gating mechanism~\cite{shrivastava2016beyond,Islam_2017_CVPR,li2019gated} that is \emph{top down} wherein larger convolutional windows and more discriminative features play a role in guiding feedforward activation among earlier features that are more local and ambiguous with respect to category.
\begin{figure}
	\begin{center}
		\setlength\tabcolsep{0.6pt}
		\def\arraystretch{0.1}	
		\resizebox{0.48\textwidth}{!}{
			\begin{tabular}{*{5}{c }}

				\includegraphics[width=0.12\textwidth]{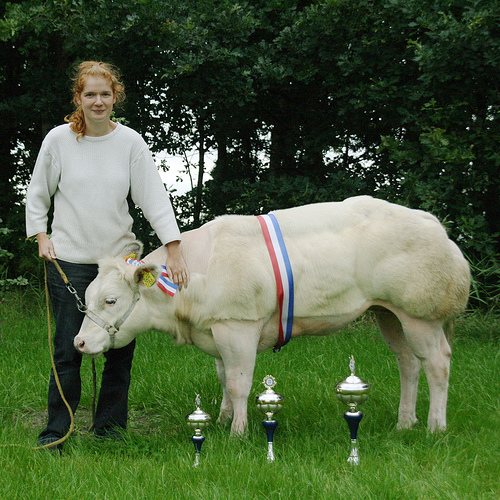}&
				\includegraphics[width=0.12\textwidth]{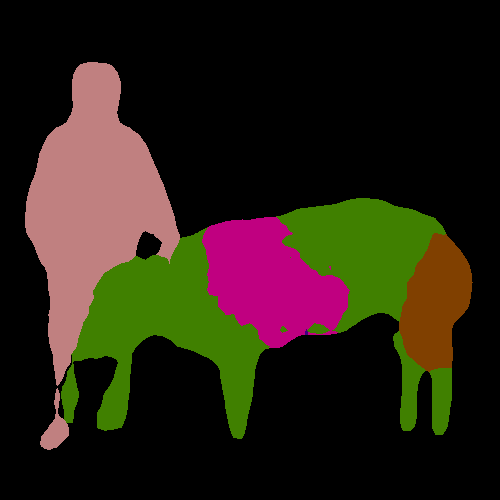}&
				\includegraphics[width=0.12\textwidth]{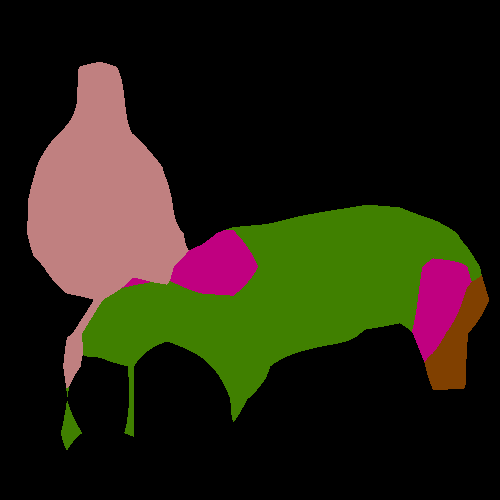}&
				\includegraphics[width=0.12\textwidth]{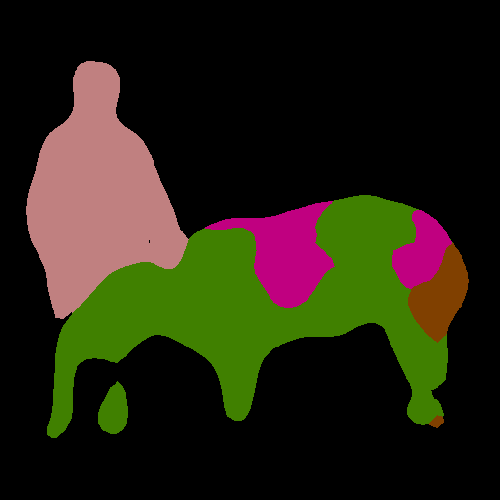}&
				\includegraphics[width=0.12\textwidth]{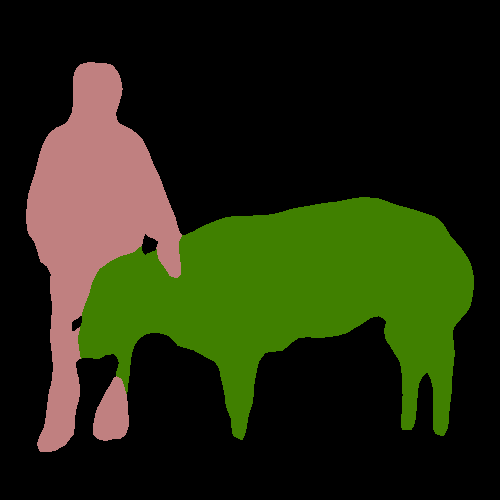}\\
				
					\includegraphics[width=0.12\textwidth]{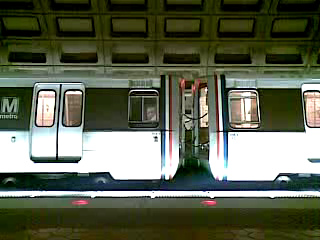}&
					\includegraphics[width=0.12\textwidth]{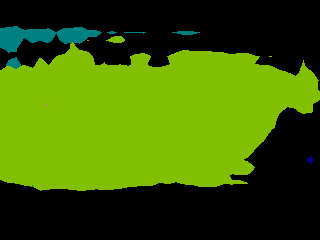}&
					\includegraphics[width=0.12\textwidth]{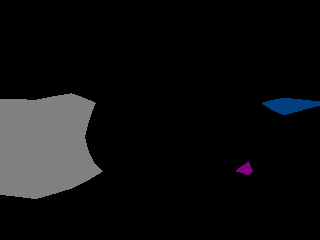}&
					\includegraphics[width=0.12\textwidth]{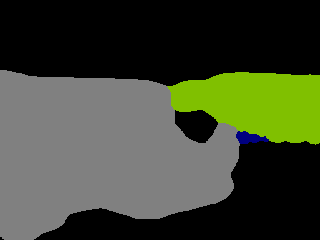}&
					\includegraphics[width=0.12\textwidth]{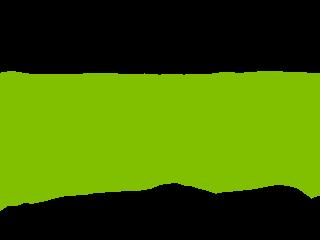}\\
				
				\includegraphics[width=0.12\textwidth]{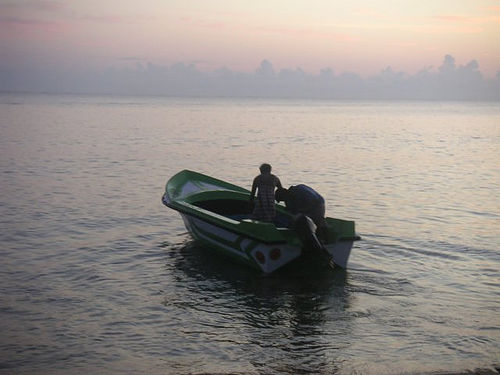}&
				\includegraphics[width=0.12\textwidth]{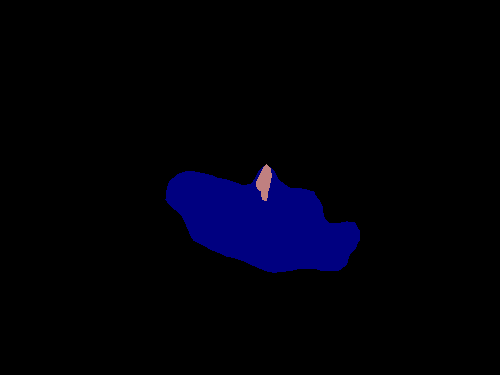}&
				\includegraphics[width=0.12\textwidth]{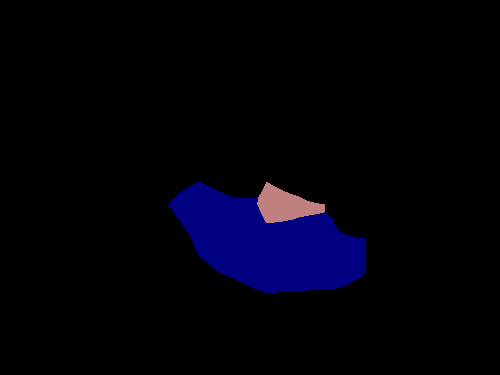}&
				\includegraphics[width=0.12\textwidth]{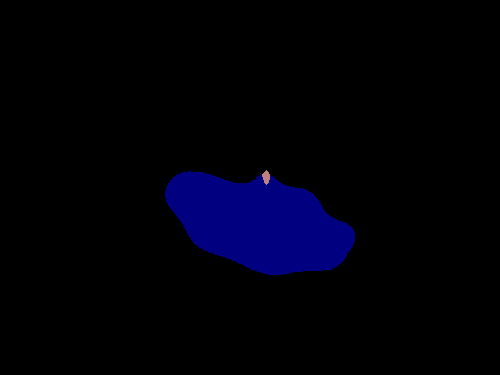}&
				\includegraphics[width=0.12\textwidth]{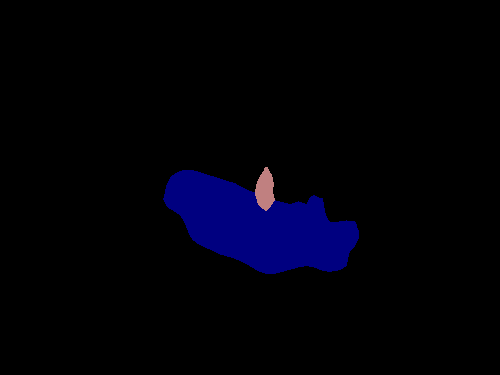}\\

				Image & ResNet101 & Stage-wise & Long-range & DIGNet \\
				\Tstrut
				
			\end{tabular}}
			\caption{ Examples of \textbf{DIGNet} predictions compared to other feedback routing mechanisms (ResNet101-8s as backbone) on PASCAL VOC 2012 dataset. Stage-wise feedback uses recurrent gating similar to~\cite{rignet}, long-range uses the initial prediction as feedback signal similar to~\cite{li2018learning,jin2017multi}, and our DIGNet uses cascaded feedback generation using propagator gates. Both stage-wise and long-range feedback fails to resolve categorical ambiguity, recover spatial details ($\mathbf{1^{st}}$ and $\mathbf{2^{nd}}$ row), and precisely segment smaller object ($\mathbf{3^{rd}}$ row)  whereas DIGNet iteratively improves predictions by refining spatial detail and diminishing representational ambiguity within the network.}
			\label{fig:intro}
		\end{center}
		\vspace{-0.6cm}
	\end{figure}
(1) The trade-off between spatial resolution for additional feature layers deeper in the network can imply a loss of spatial granularity in categorical labeling. While a simple labeling problem might be by and large globally consistent, there may remain local inconsistencies that come from this limitation. (2) The nature of convolution implies that a network is limited at any layer in the spatial extent of pixels or features that can be considered in concert or related to one another. It is also the case that the effective spatial extent of convolution among deeper layers covers a broader spatial extent by virtue of downsampling. This implies that a recurrent signal for gating can allow spatially distal discriminative features to have mutual influence over each other allowing for links to be formed between object parts that make up the whole object. (3) For intermediate features, some of these may be discriminative along certain categorical boundaries but not others with respect to the label space. An implication of this is that subject to an initial feedforward pass, features in intermediate layers may carry categorical ambiguity that is absent among more discriminative features present in deeper layers. Allowing such information to be relayed in a reverse direction can help to resolve such interference. 


The specific structure we adopt is based on a symbiotic combination of propagator and modulator nodes \textcolor{black}{which are} very flexible and highly efficient with respect to allowing information represented in one part of the network to reach other layers. The propagator gates are responsible for generating feedback signals in a cascaded manner for distributed iterative gating and modulator gates are responsible for contextual feature reweighting on selected intermediate stages based on feedback signals. This strategy is marked by a carefully designed structure for connectivity that allows for a high degree of interaction among gating and inference blocks. Moreover, the iterative (recurrent) nature of this mechanism allows for the output and internal representations to be gradually refined and also to propagate outward spatially producing an unambiguous and globally consistent prediction (see Fig.~\ref{fig:intro}).

This approach is shown to significantly boost the performance of feed-forward baselines and generate better segmentation compared to both baselines and other recurrent feedback based approaches. Furthermore, iterative inference by means of DIG is shown to converge very fast relative to other feedback mechanisms.

%% file: background.tex
\section{Related Work}\label{sec:background}
Recent state-of-the-art semantic segmentation networks~\cite{long15_cvpr,chen15_iclr,noh15_iccv,badrinarayanan15_arxiv,ghiasi2016laplacian,Islam_2017_CVPR,refinenet,chen2018deeplab} typically follow the structure of a Fully Convolutional Network (FCN). Although the feature maps produced in the higher-layers of conventional CNNs~\cite{krizhevsky12_nips,simonyan15_iclr,szegedy2015going,he2016deep} carry a strong representation of semantics, the ability to retain precise spatial details in dense labeling problems (e.g. semantic segmentation) is limited due to the poor spatial resolution.

Recent works on semantic segmentation have mainly focused on improving network performance by modifying the network architecture.
However, there are limits to the degree of improvement that is possible if the networks are confined to carry out computation based only on a single feed-forward pass. A few efforts~\cite{zamir2017feedback,mcintosh2018recurrent,pinheiro2014recurrent,rignet,kim2017iterative,li2018learning,li2016iterative,veit2017convolutional} have been proposed to iteratively improve the output of a feed-forward network and overall performance. In this work, we argue that the recurrent processing of inputs with an efficient feedback mechanism has more desirable properties, the value of which are evident in the similar mechanisms of processing observed in the human brain~\cite{gilbert2013top,lamme2000distinct}.

Several works consider employing recurrent processing~\cite{pinheiro2014recurrent, liang2015convolutional,wei2017object,kong2018recurrent,jin2017multi} or feedback based attention mechanisms~\cite{li2018learning} in combination with conventional CNNs. Another line of work~\cite{mcintosh2018recurrent,zamir2017feedback} applies a recurrent module (e.g. ConvLSTM) on top of the network to iteratively refine the initial prediction. Although feed-forward gating mechanisms~\cite{Islam_2017_CVPR} have shown some success for recognition tasks, recurrent feedback mechanisms play an important role in pushing performance further for several tasks of interest ~\cite{zamir2017feedback,carreira2016human,belagiannis2017recurrent}.

Related to our proposed approach is the idea of learning a feed-forward network in a iterative manner that involves propagating feedback in a top-down fashion. Recent feedback based approaches~\cite{zamir2017feedback,li2018learning,shrivastava2016beyond} follow the pipeline of correcting an initial prediction by propagating feedback in a few different ways. TDM~\cite{shrivastava2016beyond} proposed a pipeline where a top-down modulation network is integrated with the bottom-up feed-forward network for object detection similar to refinement based encoder-decoder architectures~\cite{refinenet,noh15_iccv,Islam_2017_CVPR,pinheiro2016learning}.

Our proposed approach differs from the above feedback based networks in that we propagate the feedback in a top-down fashion starting with the output (initial prediction) of the last layer. Our feedback mechanism iteratively adjusts the feature maps in earlier layers through feedback from higher layers and corrects initial errors towards assignment of the true category.

In summary, our feedback mechanism guides earlier features based on the feedback signal which has information from the layer immediately above, and by virtue of connectivity, from all layers above. Additionally, the iterative nature allows the feedback mechanism to carry information in a path similar to a compact hypercolumn representation and improve the quality of predictions in subsequent iterations.

%% file: approach.tex
\section{Distributed Iterative Gating Network}\label{sec:approach}
\begin{figure*}[t]
	\begin{center}
		\includegraphics[width=0.99\textwidth]{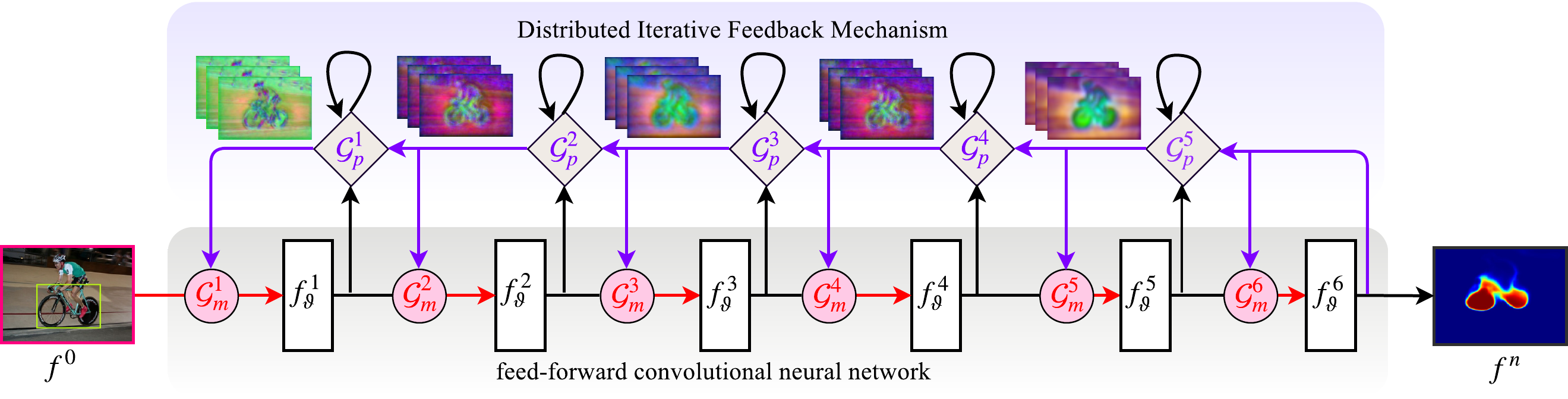}
		\caption{An illustration of our proposed \textbf{Distributed Iterative Gating Network} (DIGNet). DIGNet involves augmentation of a canonical neural network backbone through addition of gating modules, while operating in a recurrent iterative manner. ($f_\vartheta^1 \cdots f_\vartheta^6$) are bottom-up feature blocks, ($\mathcal{G}_p^1 \cdots \mathcal{G}_p^5$) are the propagator modules that propagate high-level information as feedback via a top-down pathway in order to guide the representation carried by intermediate and low-level feature layers. ($\mathcal{G}_m^1 \cdots \mathcal{G}_m^6$) are modulator gates that modulate the bottom-up flow of activation with guidance from the propagator gates. A detailed description of each component is presented in Sec.~\ref{sec:gate_modules}.}
		\label{fig:architecture}
	\end{center}
	\vspace{-0.5cm}
\end{figure*}
In this paper, we primarily focus on efficient feedback mechanisms coupled with feed-forward semantic segmentation frameworks~\cite{he2016deep,chen2018deeplab}. In general, networks that include a feedback component~\cite{li2016iterative,huang2016scene,zamir2017feedback,rignet,li2018learning} have a standard feedforward structure that consists of shallow high-resolution early spatial layers, and increasingly lower resolution richer features within deeper layers. A feedback mechanism is typically applied with iterative inference where feedback works as a correcting signal to guide the features in earlier layers based on high-level semantic representations.

Core functional parts of the feedback mechanism include (a) a selection of early or intermediate layers where a correcting signal is fed back, (b) generation of a feedback signal for each selected layer as a function of some specific deeper layers and (c) a mechanism to modulate earlier layer features applying the feedback signal. Selection of several intermediate stages~\cite{rignet,jin2017multi,li2018learning} have been found to be more effective than a recurrence mechanism that only feeds back information to the first or input layer~\cite{pinheiro2014recurrent}. Generating feedback using information from the output layer~\cite{jin2017multi,li2018learning} or from the output of deeper intermediate stages~\cite{rignet} can improve performance over baselines albeit with several limitations (See Fig.~\ref{fig:intro}). Modulating intermediate features by applying a feedback signal is generally done with additive combination or multiplicative re-weighting. In this work, we mainly focus on the second part and propose a cascaded feedback generation method that works in a distributed manner.

We have made the case that the effectiveness of a feedback mechanism may depend on considerations that include the large difference in semantically relevant or category specific representation between early and deep feature layers, or equally, the large difference in spatial resolution and spatial extent of filters typical of such networks. On one extreme, low-level features are likely to capture only concepts such as edges, contours or lines. Intuitively, allowing high-level (deep) features to directly guide low-level representations may be misguided in the absence of a satisfactory bridge provided by intermediate features in providing semantic guidance to exert influence over low-level features.  It is evident that central to the \emph{right} mechanism, is efficient integration of low and high-level features to exact all of the advantages that derive from access to both strong representation of spatial resolution, and semantically rich categorical information in a compact representation that does not introduce redundancy among features.

In the following subsections, we propose a new architecture called \emph{Distributed Iterative Gating Network} (DIGNet) that allows for feedback to propagate from deeper layers to earlier layers. This happens explicitly by virtue of connectivity among gating units, and implicitly based on updates to feedforward activation. We explain how such an architecture, namely one with a meaningful distributed feedback mechanism can produce more discriminative features by bridging the gaps in semantic specificity and resolution that exist between very deep and early layers in order to resolve categorical ambiguity.
\subsection{DIGNet Architecture}\label{DIGNet_architecture}
In this section, we introduce our proposed DIGNet that includes an efficient distributed feedback mechanism to bridge the gap between high-level and low-level features. The main objective of DIGNet is to propagate more semantic information into earlier features that will help to provide clues about semantic content within intermediate and earlier layers. We choose conventional feed-forward network architectures (e.g. ResNet101-FCN with stride 8) as our feed-forward backbone semantic segmentation network. Our proposed feedback mechanism augments the feedforward backbone with two different gating modules, (a \textit{propagator} and a \textit{modulator}) as illustrated in Fig.~\ref*{fig:architecture} to facilitate a broad exchange of information about internal representation within the network. The propagator gates together work as a lightweight parallel network that feeds information in the backward direction and generates feedback signals at each intermediate stage carrying information from all selected deeper stages. The modulator gates which are augmented between selected intermediate stages to modulate features in the feedforward backbone. The modulation is done by multiplicative re-weighting applying some weights generated from feedback signals. Details of the modulator and propagator gates are discussed in Sec.~\ref{sec:gate_modules}.

Our approach is motivated by the capacity to propagate more discriminative and semantically relevant information towards lower-layers which can be updated based on a subset of information from each downstream intermediate stage. In subsequent iterations, all stages are effectively informed in a relevance guided fashion about the outcome of all deeper stages of inference from the previous iteration. While this mechanism seems to provide greater flexibility from an intuitive perspective, and a more efficient control structure, we also find this strategy to be more helpful empirically in providing feedback in the form of an error correcting signal that modulates earlier layers to generate more discriminative features and resolve categorical ambiguity due to spatial separation of discriminative features. 

Previous efforts focus on iterative improvement leveraging earlier layer activation based on only the current prediction~\cite{jin2017multi,li2018learning}, or feedback from the feedforward stage that immediately follows the stage where refinement is occurring~\cite{rignet}. Generating feedback from the output of the last layer may have several limitations. First, as the feedback signal has much lower resolution than the input or lower layers, while resolving categorical ambiguity, it can also be misguided in regions of sharp object boundaries. Second, if there is a missing object in the initial prediction, there is a possibility that the representation of that object may be lost at some intermediate stage. In incorporating intermediate features in feedback signal generation, these limitations may be overcome. Also, generating feedback from the feedforward stage that immediately follows has a major limitation of lacking strong semantic information in earlier stages which can somewhat be improved with combining additional feedback from the last layer~\cite{rignet}. Considering the limitations of current approaches, the proper way to generate feedback signal might be combining the best of both properties, having output and all intermediate stages to contribute in feedback generation. Therefore, we generate feedback in a cascaded manner implicitly carrying some information from all the stages in an aggregated compact representation.

Intuitively, the key idea of designing the feedback mechanism in a cascade manner (deeper $\rightarrow$ shallower) can be seen as a similar to generating a hypercolumn representation~\cite{hariharan15_cvpr} where the propagated feedback signal at an earlier stage has any necessary guidance from all subsequent processing stages to correct the initial error. Naturally, the dimensionality of the feedback signal need increase with top to bottom propagation while bringing improvement subject to a hypercolumn style representation. However, in our case, the feedback signal is subject to block-wise compression through dimensionality reduction which apparently scales down the stack of feature maps by adjusting the feedback signal based on current activations before propagating towards earlier layers. The integration of this compressive strategy allows DIGNet to produce a \emph{compact hypercolumn} representation as a feedback signal. Interestingly, we find this hypothesis efficient both in terms of computational cost and performance, as our ablation results will show.
\subsection{DIGNet Iterative Inference}\label{dignet_dataflow}
In this section, we discuss the iterative inference in the DIGNet. We are using the same notation as in Fig. \ref{fig:architecture} throughout the paper. For iterative inference the recurrence is unrolled for a certain number of iterations or time steps ($T$). During the first iteration, the modulator gates ($\mathcal{G}_m^1, \mathcal{G}_m^2, \cdots ,\mathcal{G}_m^n$) simply allow a bypass of feedforward information in a bottom-up manner and hence the network works similar to feedforward networks. The feed-forward stages ($f_\vartheta^1, f_\vartheta^2, \cdots, f_\vartheta^n$) process the input image to produce a reasonable feature representation. So, DIGNet reduces to a basic feedforward network when $T=1$.

In all the subsequent iterations, DIGNet executes two steps - (a) First, feedback signals are generated in a cascaded manner starting from the initial prediction towards the earlier layers. Note that all the propagator gates ($\mathcal{G}_p^1, \mathcal{G}_p^2, \cdots ,\mathcal{G}_p^n$) are activated in this step to facilitate feedback propagation. The initial output and intermediate features flow back through the propagator network generating feedback signals for all intermediate stages. (b) Another feedforward processing flows through the backbone network with modulator gates being activated. The modulators take signals from the propagator gates and modulate the feature representation received as input from the preceding feed-forward stage before forwarding it to next stage. This step can be seen as a traditional feed-forward network except that gating interacts with feedforward processing in effect producing adaptive features. Algorithm~\ref{alg:data_flow} describes the set of steps for iterative data flow and inference in DIGNet.

Following other feedback based approaches~\cite{belagiannis2017recurrent,li2018learning,rignet,zamir2017feedback}, we optimize DIGNet  with back-propagation through time (BPTT) by unrolling the recurrence for a certain number of time steps. To elicit a trade-off between performance and computational cost we set a value of \textit{T=2} in our experiments. Note that, DIGNet does not employ any semantic supervision of intermediate predictions and only applies a cross-entropy loss to the final prediction at stage $T$. This speaks to the efficiency of communicating information broadly across the network as a loss at the final output is sufficient to realize substantive gains and effective modulator and propagator gates across the entire network.
\begin{table}[t]
	\vspace{-0.4cm}
	\begin{algorithm}[H]
		\caption{DIGNet Data Flow and Iterative Inference}\label{prune}
		\begin{algorithmic} [1]
			
			\Function{DIGNet-DF}{$\mathcal{I}$}
			\State Initialize $f^0=\mathcal{I}$
			
			\For{  t $\leftarrow$ 1 to $T_{steps}$} \Comment unroll iteration, $T$
			\If{t $>$ 1 } \Comment propagate feedback
			
			\State $\mathcal{F}^n$ = $f^n$ 
			\For{  k $\leftarrow$ $(n-1)$ to 1} 
			\State $\mathcal{F}^k = \mathcal{G}_p^k (\mathcal{F}^{k+1}, f^k)$     \Comment $ \color{antiquefuchsia}propagator$
			
			\EndFor
			\EndIf

			\For{  i $\leftarrow$ 1 to $n$}  \Comment number of stages, $n$
			
			\State $f^{{(i-1)}^\prime}$= $\mathcal{G}_m^i (\mathcal{F}^{i}, f^{i-1})$    \Comment $ \color{antiquefuchsia}modulator$
			\State $f^i$ = $f_\vartheta^i (f^{{(i-1)}^\prime})$   \Comment bottom-up feature
			
			\EndFor
			
			\EndFor
			\State return $f^n$
			\EndFunction
		\end{algorithmic}
		\label{alg:data_flow}
	\end{algorithm}
	\vspace{-0.6cm}
\end{table}
\subsection{DIGNet Gate Modules}\label{sec:gate_modules}
Here, we discuss the careful design choices for gating modules involved in the feedback mechanism.
\vspace{-0.3cm}
\subsubsection{Propagator Gate}\label{sec:pgate}
The propagator gates allow earlier layers to obtain richer semantic information in a top-down fashion, resulting in more significant interaction between low-level concepts and high-level visual features. As shown  in Fig.~\ref{fig:architecture}, each propagator gate takes a feedback signal from the previous propagator gate and bottom-up features from the corresponding intermediate stage as input to generate a new feedback signal. Intuitively, the propagator gate learns what contextual semantic information to preserve in top-down feedback propagation. The inputs are passed through a shared series of successive operations, resulting in an updated feedback signal.
The propagator module $\mathcal{G}_p$ first applies a 3$\times$3 convolution and a ReLU non-linearity, which transforms the feedback signal input $\mathcal{F}^{(i+1)}$, bottom-up features $f^i$ to $\mathcal{F}^{{(i+1)}^\prime}$ and  $f^{i^\prime}$ respectively which have a common spatial dimensionality. The resultant feature maps are then combined through concatenation followed by a $1 \times 1$ convolution to generate the feedback signal $\mathcal{F}^{i}$ which is propagated backwards to the next top-down stage. The purpose of applying convolution on the concatenated feature map is to fuse the combined feature maps and reduce channel dimensionality to ensure a compact representation. If the spatial resolution of next top-down feature map $f^{i-1}$ is higher than the feedback signal $\mathcal{F}^{i}$ then the feedback sample is upsampled by simple bilinear interpolation to have the same resolution. These operations are summarized as follows:
\vspace{-0.2cm}
\begin{gather}
\mathcal{F}^i=\hat{y}(\mathbf{W_c}*(\underbrace{(\mathbf{W_a} *f^{i})}_\text{bottom-up feature}\oplus \underbrace{ ( \mathbf{W_b} * \mathcal{F}^{i+1})}_\text{feedback signal}))
\end{gather}
where $\ast$ and $\oplus$ denote a convolution operation and concatenation, $\hat{y}$ indicates upsampling through bilinear interpolation, and \{\textbf{$\mathbf{W_a, W_b, W_c}$}\} are trainable weights. Note that the formulation for obtaining a feedback signal is the same at each top-down stage.
\subsubsection{Modulator Gate}\label{sec:mgate}
The main task of the modulator gate is to provide assistance in generating the input for the next bottom-up (feed-forward) stage by modulating information passed forward based on the feedback signal. Intuitively, the modulator gate learns to obtain a meaningful feedback signal to modulate intermediate and low-level features. The feedback signal $\mathcal{F}^{i}$ is processed first to have the same channel dimension as $f^{(i-1)}$. Inspired by~\cite{he2015spatial,liu2015parsenet,zhao2017pyramid,chen2017rethinking}, we find that applying a global contextual prior is beneficial in generating the modulating signal as shown in Table.~\ref{tab:voc2012_val_ppm}. We first create a spatial pyramid~\cite{zhao2017pyramid} of $\mathcal{F}^{i}$ with pooling rate \{1, 3, 5, 7\}. We then concatenate the pyramid features ($\mathcal{F}_1^{i^\prime}$, $\mathcal{F}_2^{i^\prime}$, $\mathcal{F}_3^{i^\prime}$, $\mathcal{F}_4^{i^\prime}$) and  $\mathcal{F}^{i}$ to obtain the updated feedback signal $\mathcal{F}^{i^\prime}$. A 1$\times$1 convolution followed by a sigmoid is applied sequentially to transform and squash the channel dimension of $\mathcal{F}^{i^\prime}$ similar to $f^{(i-1)}$, resulting in the modulating signal $\mathcal{F}^s$. Finally, $\mathcal{F}^s$ is combined with $f^{(i-1)}$ through element-wise multiplication. This new modulated bottom-up feature map $f^{(i-1)^\prime}$ passed onto the next feed-forward stage $f_\vartheta^i$ as input. 
\begin{table}[h]
	\vspace{-0.2cm}
	\begin{center}
		\def\arraystretch{1.0}
		\resizebox{0.48\textwidth}{!}{
			\begin{tabular}{c|c|cc}
				\specialrule{1.2pt}{1pt}{1pt}
				
				\multirow{2}{*}{\textbf{DIGNet-ResNet101}}	&$\ast$ & 1$\times$1 & Spatial Pyramid  \\
				
				\cline{2-4}
				
				&mIoU(\%)& 75.9 &  77.5\\
				
				\specialrule{1.2pt}{1pt}{1pt}
			\end{tabular}}
			\caption{Performance comparison of DIGNet(T=2) subject to the modulator design choices on the PASCAL VOC 2012 val set.}
			\label{tab:voc2012_val_ppm}
		\end{center}
		\vspace{-0.6cm}
\end{table}

%% file: experiments.tex
	\section{Experiments}\label{sec:exp}
To show the effectiveness of DIGNet, we present results from a series of experiments. Initially, we conduct ablation analysis to examine the impact of various design choices for DIGNet in considering the PASCAL VOC 2012 dataset~\cite{everingham2015pascal}. Then, we evaluate DIGNet on three different semantic segmentation datasets, including PASCAL VOC 2012~\cite{everingham2015pascal}, ADE20K~\cite{zhou2017scene}, and COCO-Stuff~\cite{caesar2018coco}. Experimental results demonstrate the superiority of our proposed DIGNet architecture over baselines in a variety of respects.
\subsection{Implementation Details} 
Inspired by previous work~\cite{chen2018deeplab,Islam_2017_CVPR,chen2018deeplab} we employ the \enquote{poly} learning rate policy to train the baseline networks and our DIGNet variant of the models. We employ a crop size of 321 $\times$ 321 and 513 $\times$ 513 during training and testing respectively to report experimental results on all datasets. We report experimental results for our baselines (ResNet101(32s), ResNet101(8s), and DeepLabv2-Res101) and corresponding DIGNet networks. For fairness, we use similar hyper-parameters for the baselines and our approach. We initialized baselines and our models with the COCO pre-trained weights where required, otherwise we initialize the network with ImageNet trained weights. 
Note that whenever we report experimental results for DIGNet this denotes ResNet101-DIGNet with unroll iteration, \textit{T=2}.
%
%


\subsection{Gating Semantic Information with DIGNet}
To investigate the role of distributed iterative gating in DIGNet we conduct experiments under a few different settings. We focus on two major facts to validate the hypothesis of design choices, including the significance of an iterative solution and propagating more semantic information to earlier layers by applying gating modules. 

\noindent \textbf{Iterative solution with Cascaded Feedback Generation:} In Table~\ref{tab:voc2012_val_iter}, we present quantitative results comparing different feedback routing mechanisms and significance of our iterative solution. The segmentation performance increases by a significant margin with generating feedback in a cascaded manner compared to stage-wise recurrence or with feedback from the last stage only. 
This implies that DIGNet is successful in its objectives of bridging the information gaps through efficient and effective feedback propagation.
\begin{table}[h]
	\vspace{-0.2cm}
	\begin{center}
		\def\arraystretch{1.0}
		\resizebox{0.48\textwidth}{!}{
			\begin{tabular}{c|c c c c}
				\specialrule{1.2pt}{1pt}{1pt}
				Feedback	 Method &T=1 & T=2 & T=3 & T=4\\
				\specialrule{1.2pt}{1pt}{1pt}
				Stage wise feedback~\cite{rignet} &71.3 & 73.4 &73.9 &74.9 \\
				Last layer feedback$^\dagger$~\cite{jin2017multi,li2018learning}& 71.3 & 75.1 & 74.9 & 74.5 \\
				\textbf{DIGNet} & 71.3 & 77.5 & \textbf{77.7} & 76.7 \\
				\specialrule{1.2pt}{1pt}{1pt}
			\end{tabular}}
			\caption{Performance comparison of different feedback mechanisms subject to a varying number of unrolling iterations on the PASCAL VOC 2012 val set. In all cases, T=1 reduces the network to feedforward baseline ResNet101-8s~\cite{he2016deep}.}
			\label{tab:voc2012_val_iter}
		\end{center}
		\vspace{-0.6cm}
\end{table}

Furthermore, the iterative nature of DIGNet provides adjustments to the earlier layers allowing for stage-wise feedback refinement and removing the ambiguity that may arise anywhere in the feed-forward network. We notice in Table.~\ref{tab:voc2012_val_iter} that overall performance progressively improves beyond a fixed number of iterations and then starts saturating. We find this observation to be valid across different datasets and network architectures revealing fast convergence and stability. Therefore, DIGNet may be evaluated with an increasing number of iterations to improve predictive segmentation performance.

\noindent \textbf{Semantic Information in Gating Low-level Features:} Our solution of incorporating distributed gating modules in the feedback mechanism is inspired by the following: Feed-forward network activations closer to semantic supervision tend to capture more semantics, which can guide lower-level features to correct initial errors made in inference. Instead of immediately making a category-specific prediction based on the predicted probability in the first pass, we deploy a distributed gated feedback mechanism to propagate the predicted probability to the earlier layers to update the network. In DIGNet, semantic features extracted from the last layer are passed backward as feedback which is gated with the encoded features from each stage.

\begin{table}[h]
	\vspace{-0.15cm}
	\begin{center}
		\def\arraystretch{1.0}
		\resizebox{0.45\textwidth}{!}{
			\begin{tabular}{c|cccccc|c}
				\specialrule{1.2pt}{1pt}{1pt}
				Method& $\mathcal{F}^1$ & $\mathcal{F}^2$ & $\mathcal{F}^3$ & $\mathcal{F}^4$ & $\mathcal{F}^5$  & $\mathcal{F}^6$ & mIoU(\%)    \\
				\hline
				\hline	
				
				\multirow{7}{*}{ \textbf{ResNet101-FCN}}	 & &&&&&& 65.3\\
				&    &&&&&  \checkmark & 70.5\\
				&   &&&& \checkmark& \checkmark & 71.1\\
				&    &  && \checkmark& \checkmark & \checkmark& 71.8\\
				&   & & \checkmark& \checkmark & \checkmark&  \checkmark&71.9 \\
				&  & \checkmark & \checkmark& \checkmark & \checkmark& \checkmark & 72.6\\
				&    \checkmark & \checkmark &  \checkmark &\checkmark& \checkmark & \checkmark &  \color{antiquefuchsia}\textbf{72.5} \\
				
				\specialrule{1.2pt}{1pt}{1pt}
			\end{tabular}}
			\caption{Performance of DIGNet(T=2) with a varying extent of the reach of feedback gating for the PASCAL VOC 2012 val set. }
			\label{tab:voc12_abl}
		\end{center}
		\vspace{-0.6cm}
	\end{table}
We perform a series of experiments to examine the impact of distributed gating in each feed-forward stage by selecting a subset of inferential feature blocks that are subject to gating and use them to retrain DIGNet. Experimental results are shown in Table~\ref{tab:voc12_abl}. It is clear that the segmentation quality gradually improves with the integration of more feedback propagation including to the early layers. Empirical results show that inclusion of all layers except for the initial layer sometimes achieves better results (Table 2), but inclusion of all layers in the gating process is often preferred as is the case in Table~\ref{tab:voc12_abl} and other results.
\subsection{Results on PASCAL VOC 2012 dataset}
PASCAL VOC 2012 is a popular semantic segmentation dataset consisting of 1,464 images for training, 1,449 images for validation and 1,456 images for testing, which includes 20 object categories and one background class. Following prior work~\cite{chen2018deeplab,long15_cvpr,Islam_2017_CVPR,refinenet,chen2018deeplab}, we use the augmented training set that includes 10,582 images~\cite{hariharan2011semantic}. First, we report experimental results on the PASCAL VOC 2012 validation set. We integrate DIG with ResNet-101 and Deeplabv2-ResNet101 architectures and explore the influence of the distributed feedback representation relative to the base network. Table~\ref{tab:voc2012_val_fcnresnet} shows the comparison of different baselines and our proposed approach on the PASCAL VOC 2012 validation set.
\begin{table}[ht]
	\vspace{-0.1cm}
	\begin{center}
		\def\arraystretch{1.1}
		\resizebox{0.48\textwidth}{!}{
			\begin{tabular}{rc|rc}
				\specialrule{1.2pt}{1pt}{1pt}
				\multicolumn{1}{c}{Method}&  mIoU & \multicolumn{1}{c}{Method} & mIoU  \\
				\specialrule{1.2pt}{1pt}{1pt}
				
				ResNet50-32s$^\dagger$~\cite{he2016deep}  &59.4 & \textbf{ResNet50-DIGNet} & \color{antiquefuchsia} \textbf{68} \\
			
				ResNet101-32s$^\dagger$~\cite{he2016deep} &65.3 & \textbf{ResNet101-DIGNet}&  \color{antiquefuchsia}\textbf{72.5} \\
				
				ResNet101-8s$^\dagger$~\cite{he2016deep}  &71.3  & 	\textbf{ResNet101-DIGNet} & \color{antiquefuchsia} \textbf{77.5}  \\
				
				DeepLabV2-Res101$^\dagger$~\cite{chen2018deeplab}  & 74.9 &  \textbf{DeepLabV2-DIGNet} & \color{antiquefuchsia}\textbf{76.1}\\

				\specialrule{1.2pt}{1pt}{1pt}
			\end{tabular}}
			\caption{PASCAL VOC 2012 validation set results for baselines and DIGNet(T=2).}
			\label{tab:voc2012_val_fcnresnet}
		\end{center}
		\vspace{-0.4cm}
\end{table}
Interestingly, \emph{ResNet101-DIGNet} with \textit{OS=32} marginally outperforms ResNet101-FCN with \textit{OS=8} in terms of mIoU achieving 72.5\% and 71.3\% respectively. Also, \emph{ResNet101-DIGNet} with \textit{OS=8} yields better performance than Deeplabv2-ResNet101 providing a strong case for the value of our proposed distributed iterative feedback mechanism. Additionally, \emph{DeeplabV2-DIGNet} significantly outperforms the baseline and achieves 76.1\% mIoU without any \emph{bells and whistles}. It is observed that the performance consistently increases for the baselines with the addition of DIG.

We further conduct experiments for the proposed DIGNet on the PASCAL VOC 2012 test set. Following existing works~\cite{chen2018deeplab,zhao2017pyramid,noh15_iccv,refinenet}, DIGNet is first trained on the augmented training set and then fine-tuned on the original PASCAL VOC 2012 trainval set. We evaluate DIGNet with muti-scale inputs including left-right flips, where the scales are \{0.5, 0.75, 1.0, 1.25, 1.5\}, and average the multi-scale outputs for final predictions. As shown in Table~\ref{tab:quant_pascal}, DIGNet achieves 80.7\% mIoU which is competitive compared to other baselines especially for a simple mechanism attached to a standard ResNet architecture. Note that, unlike many recent works, we did not employ hardware intensive optimization like training batch norm parameters , extremely time consuming procedures like pre-training on large scale databases with semantic supervision, or combining multiple loss functions to boost performance yet we observe dramatic performance gains.
\begin{table}[t]
	\begin{center}
		\def\arraystretch{1.0}
		\resizebox{0.38\textwidth}{!}{
			\begin{tabular}{l|c}
				\specialrule{1.2pt}{1pt}{1pt}
				\multicolumn{1}{c}{Method} &  mIoU (\%)\\
				\specialrule{1.2pt}{1pt}{1pt}
				Adelaide\_Very\_Deep\_FCN\_VOC~\cite{wu2016bridging} & 79.1 \\
				
				LRR\_4x\_ResNet-CRF~\cite{ghiasi2016laplacian}&79.3\\
				DeepLabv2-CRF~\cite{chen2018deeplab} & 79.7\\
				CentraleSupelec Deep G-CRF~\cite{chandra2016fast} & 80.2\\
				SegModel~\cite{shen2017semantic} & 81.8\\
				Deep Layer Cascade (LC)~\cite{li2017not}&82.7\\
				TuSimple~\cite{wang2018understanding} &83.1\\
				Large\_Kernel\_Matters~\cite{peng2017large}&83.6\\
				Multipath-RefineNet (Res152)~\cite{refinenet} & 83.4\\
				PSPNet~\cite{zhao2017pyramid} &85.4\\
				DeepLabv3~\cite{chen2018deeplab} & 85.7 \\
				\midrule
				\textbf{DIGNet} & 80.7 \\	
				
				\specialrule{1.2pt}{1pt}{1pt}
			\end{tabular}}
			\caption{Quantitative results in terms of mean IoU on PASCAL VOC 2012 test set.}
			\label{tab:quant_pascal}
		\end{center}
		\vspace{-0.5cm}
\end{table}

\begin{figure}[h]
	\vspace{-0.1cm}
	\begin{center}
		\setlength\tabcolsep{0.7pt}
		\def\arraystretch{0.5}
		\resizebox{0.47\textwidth}{!}{
			\begin{tabular}{*{5}{c }}
				
%
				
%

				%
				\includegraphics[width=0.2\textwidth]{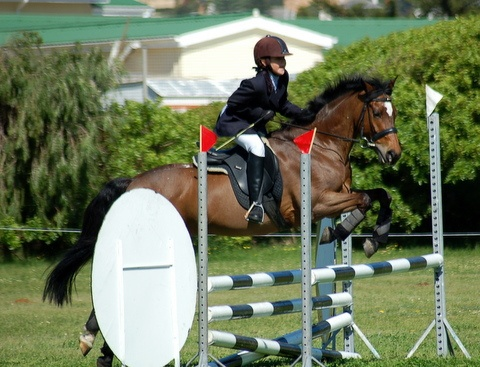}&
				\includegraphics[width=0.2\textwidth]{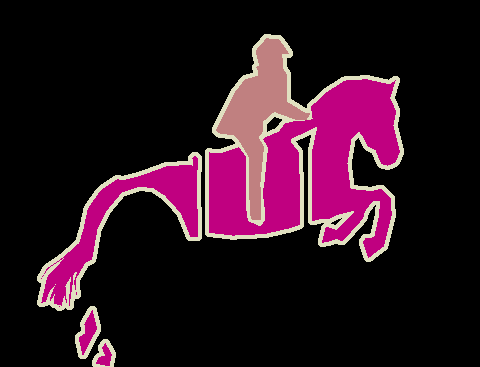}&
				\includegraphics[width=0.2\textwidth]{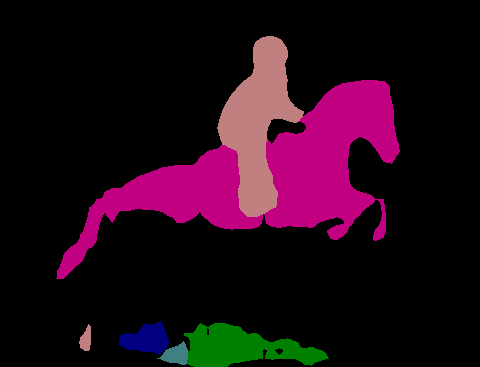}&
				\includegraphics[width=0.2\textwidth]{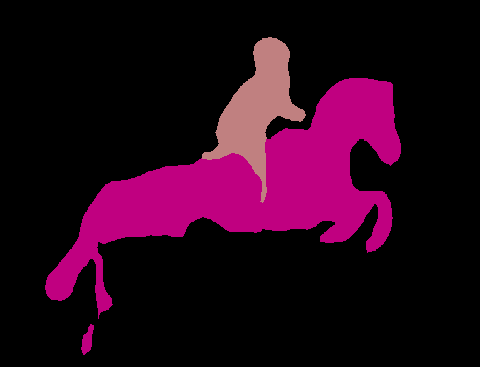}\\

								\includegraphics[width=0.2\textwidth]{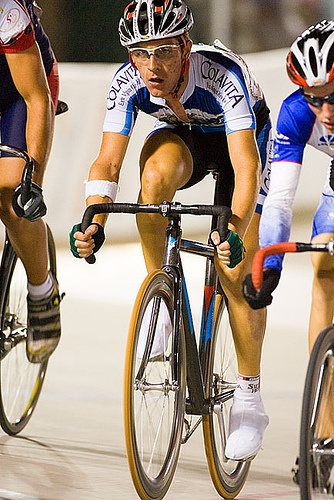}&
								\includegraphics[width=0.2\textwidth]{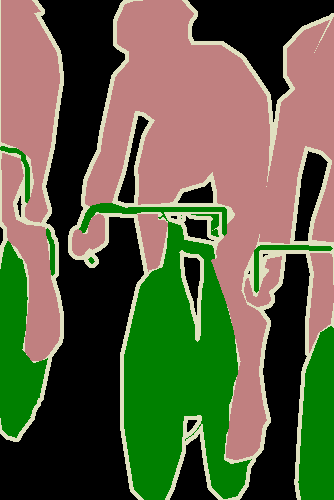}&
								\includegraphics[width=0.2\textwidth]{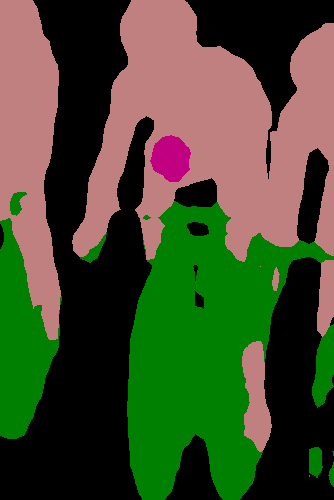}&
								\includegraphics[width=0.2\textwidth]{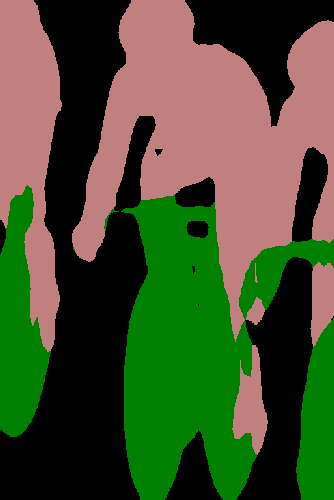}\\

				Image & ground-truth& ResNet101 & \textbf{DIGNet} \Tstrut
			\end{tabular}}
			\caption{Qualitative results of DIGNet corresponding to the PASCAL VOC 2012 validation set.}
			\label{fig:val_pascal}
		\end{center}
		\vspace{-0.5cm}
\end{figure}

We provide a qualitative visual comparison of our approach with respect to the baselines in Fig.~\ref{fig:val_pascal}. With the proposed mechanism, we produce improved prediction results compared to the baselines and many of these regions are re-examined and refined with the help of DIG.

\subsection{Results on ADE20K}
ADE20K is a newer and more complex dataset for scene parsing that provides semantic labels for 150 classes including 115 thing categories and 35 stuff categories, with more than 20k indoor and outdoor images. Table~\ref{tab:ade20k} presents the scene parsing results obtained with the ADE20K validation set for different baselines and our proposed approach. With ResNet101(8s) \emph{DIGNet} alone yields 36.9\% mIoU, significantly outperforming ResNet101-FCN and DeepLab-Res101 by about 3.3\% and 1.6\%, respectively. Additionally, DeepLabv2-DIGNet achieves 36.9\% mIoU which outperforms the baseline.
\begin{table}[h]
	\begin{center}
		\def\arraystretch{1.0}
		\resizebox{0.48\textwidth}{!}{
			\begin{tabular}{l|ccc}
				\specialrule{1.2pt}{1pt}{1pt}
				Method& mIoU(\%)  & Pixel Acc.(\%) & Overall(\%)   \\
				\specialrule{1.2pt}{1pt}{1pt}
				CascadeNet~\cite{zhou2017scene} &34.90 & 74.52 &54.71 \\
				DilatedNet~\cite{yu2015multi}  &34.3 & 76.4 & 55.3 \\
				PSPNet~\cite{zhao2017pyramid} &41.7 & 80.0 & 60.9 \\
				\midrule
				
				ResNet101$^\dagger$ & 33.6 & 75.4 & 44.2 \\
				\textbf{ResNet101-DIGNet} & 36.9 & 77.3& 46.6\\
				
				\midrule
				DeepLabv2-ResNet101$^\dagger$ & 35.3 & 75.5 &  45.1 \\
				\textbf{DeepLabv2-DIGNet} & 36.9 & 76.7& 47.8\\
				

				\specialrule{1.2pt}{1pt}{1pt}
		\end{tabular}}
		\caption{ Quantitative analysis of our approach based on different architectures \textit{vs.} state-of-the-art methods based on the ADE20K validation set. $\dagger$ indicates our implementation.} 
		\label{tab:ade20k}
	\end{center}
	\vspace{-0.7cm}
\end{table}
\subsection{Results on COCO-Stuff10k}
COCO-Stuff10k is also a relatively recently released scene parsing dataset based on MS-COCO annotations. Following the split in~\cite{caesar2018coco}, we use 9k images for training and another 1k for testing to evaluate DIGNet. We further evaluate our model on the scene centric large-scale COCO-Stuff dataset to examine the value of the proposed distributed iterative gating mechanism. Comparison of scene parsing results on the COCO-Stuff dataset are reported in Table~\ref{tab:compare_cocostuff}. Similar to previous experiments, we mainly focus on the effect of augmenting ResNet based architectures using DIGNet. Augmenting ResNet101(32s) for DIGNet provides improvement of 2.7\% over the baseline. Similarly, augmenting ResNet101(8s) improves the performance significantly (33.4\% \textit{v.s.} 36.9\%). We further apply DIG on the DeepLabv2 network which improves the baseline to some degree (34.1\% \textit{v.s.} 35.9\%). For this challenging dataset, these improvements are quite significant.

\begin{table}[h]
	\vspace{-0.15cm}
	\begin{center}
		\def\arraystretch{1.0}
		\resizebox{0.44\textwidth}{!}{
			\begin{tabular}{l|ccc}
				\specialrule{1.2pt}{1pt}{1pt}
				Method& pAcc(\%) & mAcc(\%) & mIoU(\%)  \\
				\specialrule{1.2pt}{1pt}{1pt}
				DeepLab~\cite{chen15_iclr} & 57.8 &38.1&	26.9\\
				OHE + DC + FCN~\cite{hu2017labelbank} &  66.6&45.8& 34.3\\
				DAG-RNN + CRF~\cite{shuai2017scene} &   	63.0 &42.8 &	31.2 \\
				RefineNet-Res101~\cite{refinenet} & 65.2 & 45.3& 33.6 \\
				CCL~\cite{ding2018context}& 66.3 & \textbf{48.8} & 35.7 \\
				\midrule
				
				ResNet101-32s~\cite{he2016deep}$^\dagger$&  58.7 & 38.0 &26.4 \\
				 \textbf{ResNet101-DIGNet}& 61.8  & 40.7 &29.1 \\
				\midrule
				ResNet101-8s~\cite{he2016deep}$^\dagger$&  64.6 & 44.9 &33.4 \\
				\textbf{ResNet101-DIGNet}& \textbf{67.3}  & 47.4 &\textbf{36.3} \\
				\midrule
				DeepLabv2 (ResNet-101)$^\dagger$~\cite{chen2018deeplab} & 65.1&45.5& 34.1 \\
				\textbf{DeepLabv2-DIGNet} & 67.0 &  46.4 & 35.9  \\
				
				\specialrule{1.2pt}{1pt}{1pt}
		\end{tabular}}
		\caption{Comparison of scene parsing results on the Coco-Stuff test set. $\dagger$ refers to our own implementation.}
		\label{tab:compare_cocostuff}
	\end{center}
	\vspace{-0.6cm}
\end{table}
\subsection{Study of Error Correction with DIGNet}
We characterize the computational properties of generic unrolling operations in DIGNet given that it performs identical operations in each iteration. We address this consideration from three different vantage points by focusing on the initial prediction of the feed-forward network.

\begin{figure}[h]
	\begin{center}
		\setlength\tabcolsep{1.0pt}
		\def\arraystretch{0.5}
		\resizebox{0.46\textwidth}{!}{
			\begin{tabular}{*{4}{c }}
				
				\includegraphics[width=0.15\textwidth]{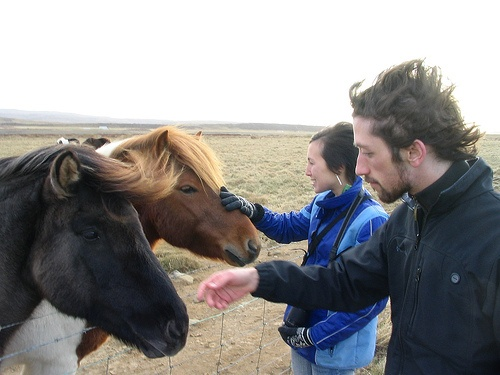}&
				\includegraphics[width=0.15\textwidth]{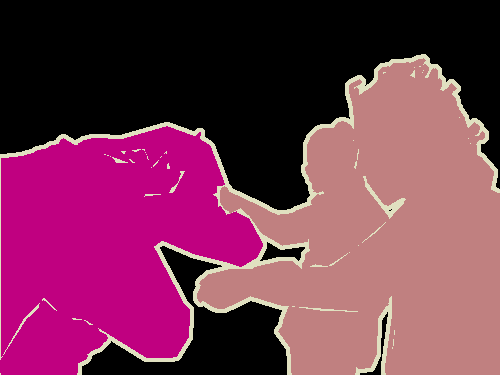}&
				\includegraphics[width=0.15\textwidth]{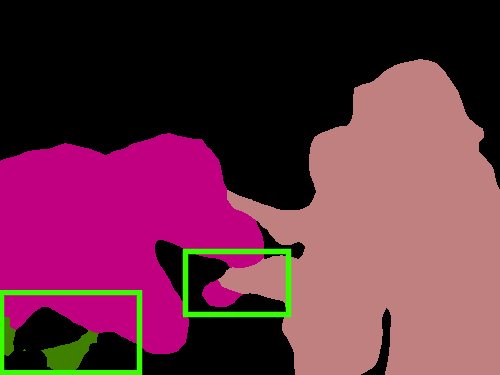}&
				\includegraphics[width=0.15\textwidth]{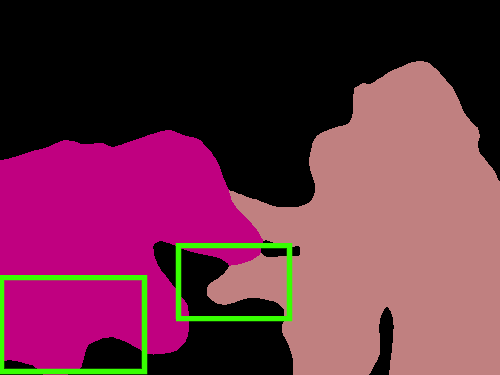}\\
				
				
				image & ground-truth & Baseline & DIGNet  \\
						
		\end{tabular}}
		\caption{When the initial prediction has categorical ambiguity, DIGNet iteratively adjusts information passed forward through the feedback signal resulting in recognition of the correct class.}
		\label{fig:discuss}
	\end{center}
    \vspace{-0.8cm}
\end{figure}

\noindent \textbf{Categorical Ambiguity:} When the initial class assignment is predicted incorrectly - for instance segmenting a horse as a cow or vice versa- we empirically found that DIGNet is capable of correcting the initial prediction in the very first iteration (see Fig.~\ref{fig:discuss}), highlighting the powerful influence of DIGNet to correct the poor initial prediction. 

\begin{figure}[h]
	\vspace{-0.5cm}
	\begin{center}
		\setlength\tabcolsep{1.1pt}
		\def\arraystretch{0.5}
		\resizebox{0.47\textwidth}{!}{
			\begin{tabular}{*{4}{c }}

				\includegraphics[width=0.15\textwidth]{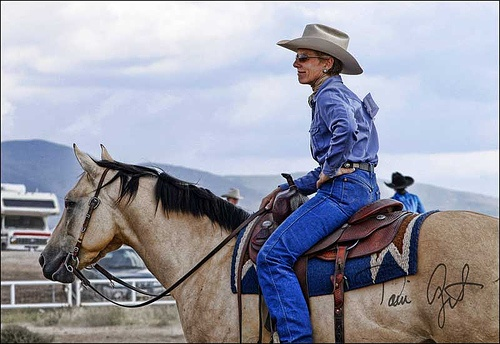}&
				\includegraphics[width=0.15\textwidth]{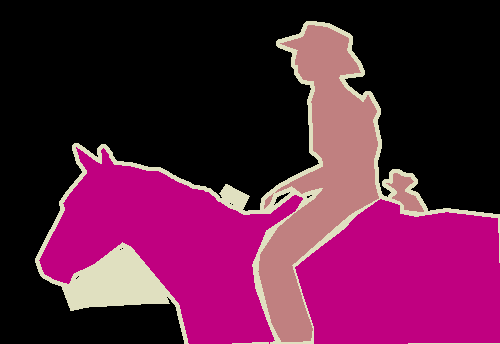}&
				\includegraphics[width=0.15\textwidth]{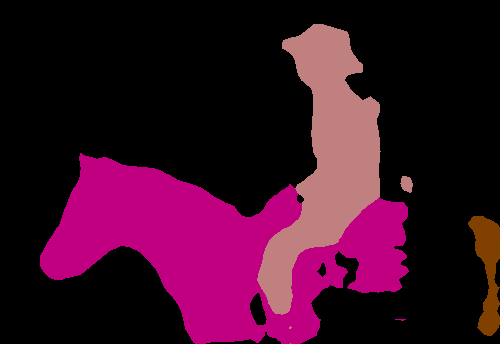}&
				\includegraphics[width=0.15\textwidth]{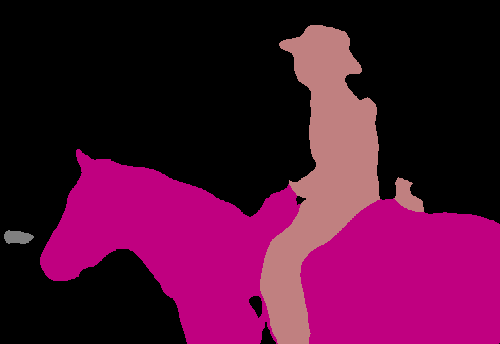}\\

				image & ground-truth & baseline & DIGNet \\

		\end{tabular}}
		\caption{When the initial prediction is able to detect a part of an object, DIGNet gradually aligns the output more accurately with semantic labels, while labeling the initially missing regions.}
		\label{fig:discuss3}
	\end{center}
    \vspace{-0.6cm}
\end{figure}

\noindent \textbf{Partial Segmentation:} When the initial prediction has coarse-grained or spatially limited mask (see Fig.~\ref{fig:discuss3}), DIGNet improves partial segmentation to generate a detailed mask by incorporating distributed gating in the feedback mechanism, in some instances completing the object.

\begin{figure}[h]
	\vspace{-0.2cm}
	\begin{center}
		\setlength\tabcolsep{1.1pt}
		\def\arraystretch{0.5}
		\resizebox{0.47\textwidth}{!}{
			\begin{tabular}{*{4}{c }}
				
				\includegraphics[width=0.2\textwidth]{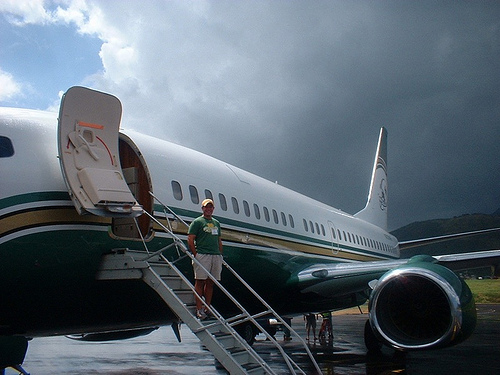}&
				\includegraphics[width=0.2\textwidth]{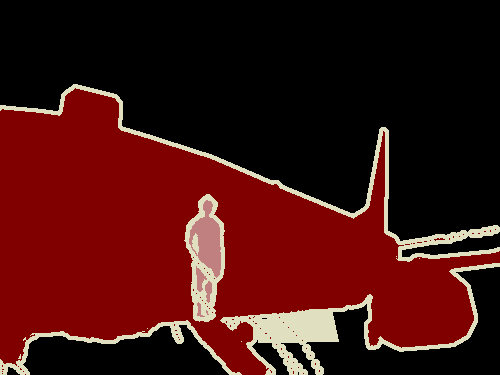}&
				\includegraphics[width=0.2\textwidth]{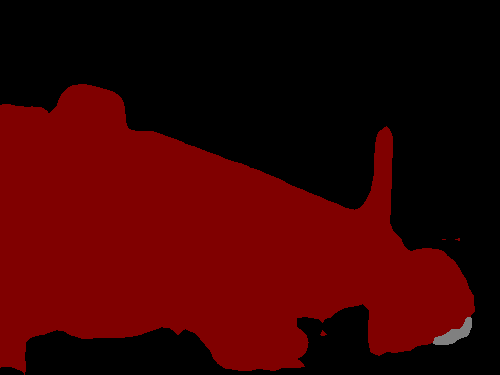}&
				\includegraphics[width=0.2\textwidth]{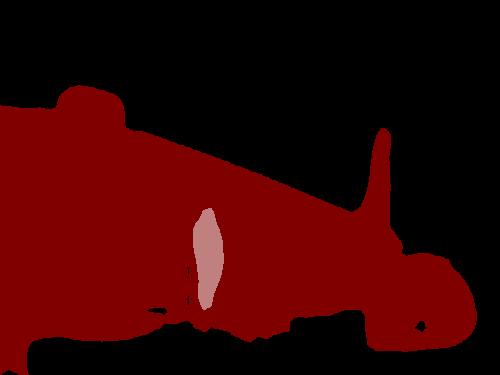}\\
				image & ground-truth & baseline & DIGNet \\

			\end{tabular}}
			\caption{When the initial prediction is able to detect a part of an object, DIGNet gradually aligns output more accurately with semantic labels, while labeling the initially missing regions.}
			\label{fig:missing_object}
		\end{center}
		\vspace{-0.6cm}
	\end{figure}

\noindent \textbf{Recover Missing Small Objects:} When the initial prediction misses small objects in front of large objects, DIGNet can recover missing small objects (see Fig.~\ref{fig:missing_object}). DIGNet succeeds because when feedback is generated in a cascaded manner including intermediate feature maps, some earlier representation where the object is more strongly represented is fed back to guide the next iteration.

\noindent \textbf{Coarse-to-Fine Representation:} DIGNet processes at a relatively coarse spatial resolution due to the output stride applied on the image features with the absence of a refinement/decoder network. While the performance improvement is remarkable in just one additional iteration with DIGNet, we show that the hierarchical addition of propagator and modulator gates in the feedback mechanism can be represented as a coarse-to-fine refinement scheme.
\begin{figure}[h]
	\vspace{-0.1cm}
	\begin{center}
		\setlength\tabcolsep{1.1pt}
		\def\arraystretch{0.5}
		\resizebox{0.48\textwidth}{!}{
			\begin{tabular}{*{6}{c }}

				\includegraphics[width=0.16\textwidth]{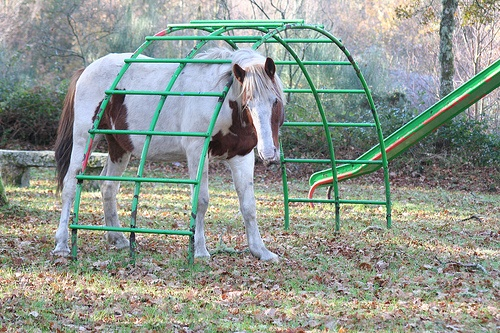}&
				\includegraphics[width=0.16\textwidth]{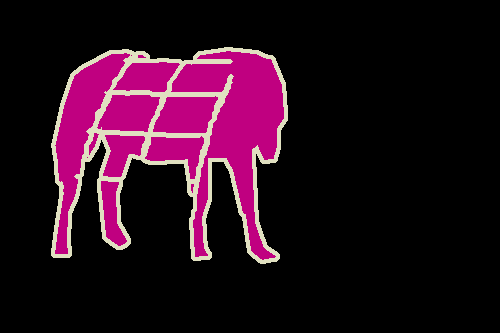}&
				\includegraphics[width=0.16\textwidth]{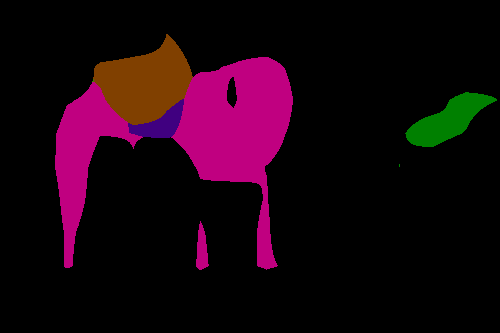}&
				\includegraphics[width=0.16\textwidth]{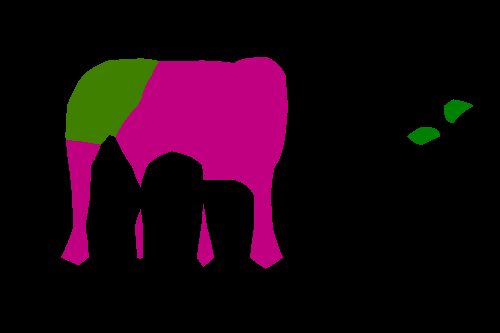}&
				\includegraphics[width=0.16\textwidth]{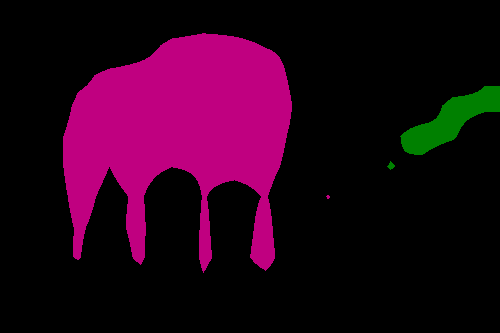}&
				\includegraphics[width=0.16\textwidth]{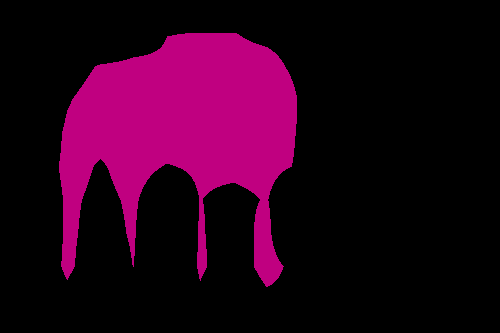} \\

				image & ground-truth&ResNet101&  DIGNet $\ll \mathcal{F}^4 \gg$   & DIGNet $\ll \mathcal{F}^2 \gg$ \Tstrut & DIGNet \\
				
		\end{tabular}}
		\caption{Visualization of label quality after top-down addition of distributed iterative gating modules. For each row, we show the input image, ground-truth, ResNet101(32s) prediction, and the predicted label map of DIGNet when distributed iterative gating modules are included in a top$\rightarrow$down manner.}
		\label{fig:discuss4}
	\end{center}
    \vspace{-0.6cm}
\end{figure}

Fig.~\ref{fig:discuss4} illustrates the degree of refinement obtained after integrating stage-wise gating modules. DIGNet$\ll\mathcal{F}^n\gg$ refers to feedback propagated until block $n$. Interestingly, with the addition of top-down recurrent feedback, DIGNet predictions continue to improve by recovering spatial details while aligning to resolve categorical ambiguity.

DIGNet's ability to iteratively resolve categorical ambiguity with precise localization of sharp object boundaries (Fig.\ref{fig:discuss}), improve partial segmentation (Fig.\ref{fig:discuss3}), and correct initial errors by way of coarse-to-fine refinement (Fig.\ref{fig:discuss4}) provides a convincing case for the effectiveness of distributed iterative gating mechanism. 
\subsection{Analyzing the Failure Cases of DIGNet}
\begin{figure}[h]
	\vspace{-0.2cm}
	\begin{center}
		\setlength\tabcolsep{1.1pt}
		\def\arraystretch{0.5}
		\resizebox{0.48\textwidth}{!}{
			\begin{tabular}{*{5}{c }}
				
				\includegraphics[width=0.15\textwidth]{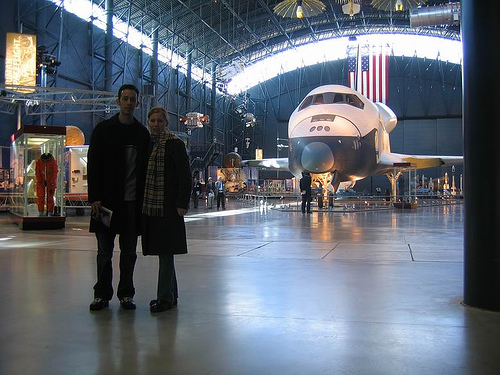}&
				\includegraphics[width=0.15\textwidth]{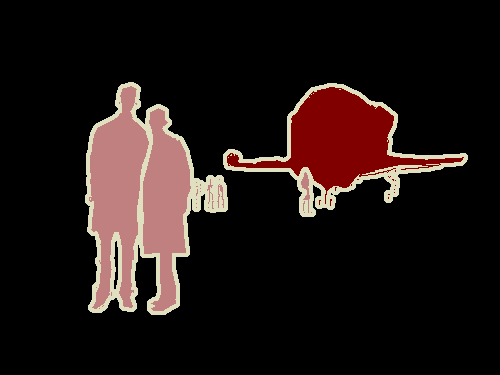}&
				\includegraphics[width=0.15\textwidth]{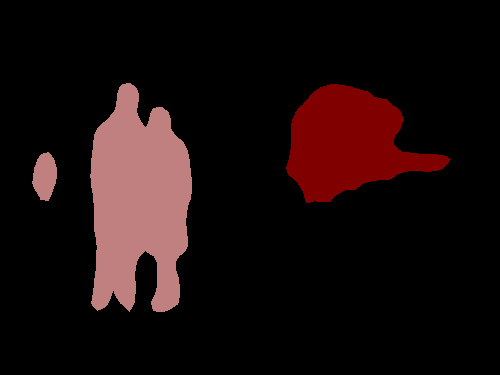}&
				\includegraphics[width=0.15\textwidth]{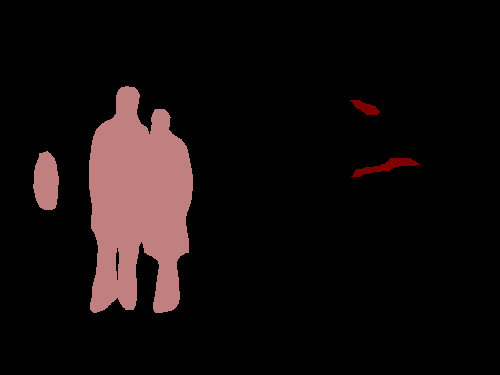}\\

				image & ground-truth & baseline & DIGNet \\

		\end{tabular}}
		\caption{An example where the final labeling in each case tends to be globally consistent over objects.} 
		\label{fig:failure}
	\end{center}
	\vspace{-0.8cm}
\end{figure}

Despite the consistent performance improvement for the majority of cases, there are rare cases that are more challenging to predict. When DIGNet is allowed to iteratively propagate high-level semantics to earlier layers, it progressively improves the label map by way of top-down modulation (Fig.~\ref{fig:discuss4} and Table~\ref{tab:voc12_abl}).
In extreme cases, when the initial prediction of any foreground object shares similar visual features with the surrounding background it may gradually move to partial incorrect labeling (see Fig.~\ref{fig:failure}). Here, DIGNet is able to predict the correct class including both the people and the airplane in the background. However, when feedback modulates the feedforward signal, the airplane is suppressed. Such a case may occur when confidence in two classes is very similar and the airplane in this case shares notable features with the background. Interestingly, the label is globally consistent which underscores the ability to successfully propagate confidence spatially despite an incorrect adjustment to the class label.

%% file: conclude.tex
\section{Conclusion}\label{sec:conclude}
In this paper we have presented a scheme for \emph{Distributed Iterative Gating} called DIGNet. This strategy involves iterative inference by unfolding a recurrent architecture for a certain number of time steps that includes a cascaded feedback signal guiding shallower layers to learn more discriminative adaptive features. This is achieved through a carefully designed top-down structure that allows all deeper layers the potential to influence feedforward inference. Ablation studies and associated analysis reveal a strong capacity for spatial and categorical ambiguity to be resolved across feature layers and over space with rapid convergence on an optimal decision. Furthermore, DIGNet presents promising potential for improving inference capability of semantic segmentation in challenging cases with a feedback guided iterative inference mechanism.